%% file: acl_latex.tex
\documentclass[11pt]{article}

\usepackage[final]{acl}
\usepackage{times}
\usepackage{latexsym}
\usepackage{booktabs}
\usepackage{multirow}
\usepackage{array}
\usepackage{graphicx} 
\usepackage{amsmath}
\usepackage{amsmath}
\usepackage{tcolorbox}
\usepackage{enumitem}
\usepackage[table]{xcolor}
\usepackage{colortbl}
\definecolor{zebra}{gray}{0.94}
\definecolor{zsBg}{RGB}{235,243,251}
\definecolor{sftBg}{RGB}{251,238,235}
\definecolor{deltaBg}{RGB}{245,245,245}
\newcommand{\zc}[1]{\cellcolor{zebra}#1}

\definecolor{deltaneg}{RGB}{183,28,28}
\definecolor{deltapos}{RGB}{27,94,32}
\newcommand{\dneg}[1]{\textcolor{deltaneg}{\scriptsize $-$#1}}
\newcommand{\dpos}[1]{\textcolor{deltapos}{\scriptsize $+$#1}}

\usepackage[T1]{fontenc}

\usepackage[utf8]{inputenc}

\usepackage{microtype}

\usepackage{inconsolata}

\usepackage{graphicx}

\title{Beyond Polarization: The Generative Constraint of Chain-of-Thought in Pointwise Reranking}

\author{
 \textbf{Xiaoyang Chen\textsuperscript{1,2,3}},
 \textbf{Jie Liu\textsuperscript{3}},
 \textbf{Haijin Liang\textsuperscript{3}},
 \textbf{Haibo Shi\textsuperscript{3}},
\\
 \textbf{Jin Ma\textsuperscript{3}},
 \textbf{Ben He\textsuperscript{1,2}},
 \textbf{Yingfei Sun\textsuperscript{1}},
 \textbf{Dezhi Ye \textsuperscript{3}\thanks{~Corresponding author.}}
\\
 \textsuperscript{1}University of Chinese Academy of Sciences, \\
 \textsuperscript{2}Chinese Information Processing Laboratory, Institute of Software, \\ Chinese Academy of Sciences,\\
 \textsuperscript{3}Tencent
\\
 \texttt{chenxiaoyang19@mails.ucas.ac.cn},
 \texttt{dezhiye@tencent.com}
}

\begin{document}
\maketitle
\begin{abstract}

 In pointwise document reranking, Chain-of-Thought models typically underperform direct scoring models. While existing diagnostics attribute this to inferior classification, score polarization, or calibration breakdown, whether targeted training can bridge this gap remains unclear. Our empirical study first confirms that this gap is stable across scales up to 32B parameters, ruling out model and data capacity confounders. We then apply stress tests utilizing reinforcement learning, fine-grained supervision, and architectural decoupling to explicitly repair these deviations. Although these interventions improve classification accuracy and absolute scores, the relative ranking gap persists.
 These findings suggest that, within the pointwise scoring paradigm, routing continuous relevance semantics through discrete text constrains ranking signal resolution, revealing a bottleneck that is stable and difficult to overcome under current standard methods, rather than an easily resolvable training bias.
\end{abstract}

\input{latex/introduction}

\input{latex/experiment}

\input{latex/stage1}

\input{latex/stage2}

\section*{Limitations}
The primary limitation of this study lies in its scope. Although we cover a range of model sizes, data scales, and CoT sources, we cannot entirely rule out all potential confounders, including specific task types, data distribution peculiarities, and training-regime factors such as rationale length and style that we cannot fully decouple within a single paper, and our 200-sample human evaluation of rationale quality offers only limited guarantees. Following standard pointwise conventions, we read the score at a fixed position, the token after \texttt{Answer:}, and we leave aside variants that alter the attention or aggregation so that the scorer attends more freely to all preceding hidden states rather than being bottlenecked by the discrete text $t$, which lies beyond the scope of a short paper and could affect the observed gap. Our stress tests target the specific hypotheses raised in prior work, namely classification accuracy, score polarization, and calibration breakdown, rather than exhaustively isolating every training variable, so repair paths such as noCoT-to-CoT distillation, continuous regression heads, and ranking-oriented losses remain uncovered and may narrow the gap further. Finally, our diagnostic perspective focuses primarily on internal model mechanisms and does not fully consider external factors such as training resource constraints or the complexities of real-world application, which may significantly affect model performance. We therefore temper our claims accordingly and leave a more complete decoupling to future work.



\bibliography{custom}

\appendix

\input{latex/appendix}

\end{document}

%% file: latex/introduction.tex
\begin{table*}[htbp!]
\centering
\setlength{\belowcaptionskip}{-0.4cm}
\resizebox{\textwidth}{!}{%
\begin{tabular}{l l cc cc cc !{\vrule width 0.6pt} ccc ccc ccc}
\toprule
\multirow{3}{*}{\textbf{Model}} & \multirow{3}{*}{\textbf{Size}}
& \multicolumn{6}{c!{\vrule width 0.6pt}}{\textbf{Zero-Shot}} 
& \multicolumn{9}{c}{\textbf{SFT}} \\
\cmidrule(lr){3-8}\cmidrule(l){9-17}
& & \multicolumn{2}{c}{BRIGHT} & \multicolumn{2}{c}{DL19} & \multicolumn{2}{c!{\vrule width 0.6pt}}{DL20}
& \multicolumn{3}{c}{BRIGHT} & \multicolumn{3}{c}{DL19} & \multicolumn{3}{c}{DL20} \\
\cmidrule(lr){3-4}\cmidrule(lr){5-6}\cmidrule(lr){7-8}
\cmidrule(lr){9-11}\cmidrule(lr){12-14}\cmidrule(lr){15-17}
& & noCoT & CoT & noCoT & CoT & noCoT & CoT
& noCoT & DS & Gem & noCoT & DS & Gem & noCoT & DS & Gem \\
\midrule

\multirow{3}{*}{Qwen2.5}
& 7B  & 11.1 & \textbf{15.3} & \textbf{61.2} & 60.0 & \textbf{55.6} & 55.0
       & \textbf{21.7} & 21.0 & 18.0 & \textbf{73.2} & 70.5 & 68.7 & \textbf{69.5} & 64.9 & 62.6 \\
& 14B & \zc{\textbf{19.7}} & \zc{17.8} & \zc{\textbf{63.4}} & \zc{63.2} & \zc{\textbf{59.1}} & \zc{57.0}
       & \zc{22.1} & \zc{\textbf{22.6}} & \zc{18.8} & \zc{\textbf{72.1}} & \zc{66.7} & \zc{67.8} & \zc{\textbf{68.7}} & \zc{65.5} & \zc{64.5} \\
& 32B & \textbf{21.3} & 17.7 & \textbf{63.5} & 62.9 & \textbf{56.7} & 54.9
       & \textbf{24.4} & 22.6 & 19.5 & \textbf{73.1} & 69.2 & 71.1 & \textbf{68.8} & 65.0 & 65.1 \\
\midrule
 
\multirow{4}{*}{Qwen3}
& 0.6B & --   & 9.2  & --   & 44.1 & --   & 38.9
        & \textbf{16.9} & 9.6  & 12.3 & \textbf{71.4} & 60.1 & 62.5 & \textbf{67.2} & 55.1 & 55.7 \\
& 8B   & \zc{\textbf{19.8}} & \zc{16.1} & \zc{55.5} & \zc{\textbf{55.6}} & \zc{45.9} & \zc{\textbf{50.3}}
        & \zc{\textbf{23.5}} & \zc{19.8} & \zc{20.3} & \zc{\textbf{71.2}} & \zc{66.4} & \zc{64.2} & \zc{\textbf{69.1}} & \zc{62.0} & \zc{57.3} \\
& 14B  & \textbf{21.2} & 18.8 & \textbf{58.4} & 58.3 & \textbf{53.8} & \textbf{53.8}
        & \textbf{22.7} & 21.1 & 19.7 & \textbf{69.9} & 66.8 & 65.7 & \textbf{68.5} & 63.3 & 61.7 \\
& 32B  & \zc{\textbf{22.4}} & \zc{17.2} & \zc{\textbf{67.7}} & \zc{58.5} & \zc{\textbf{65.5}} & \zc{52.2}
        & \zc{\textbf{24.6}} & \zc{23.0} & \zc{21.8} & \zc{\textbf{73.7}} & \zc{70.1} & \zc{67.4} & \zc{\textbf{69.9}} & \zc{67.3} & \zc{65.5} \\
\bottomrule
\end{tabular}%
}
\caption{
NDCG@10 of pointwise reranking on BRIGHT, TREC DL19\&20 across the Qwen2.5/Qwen3 series. We compare \textit{Direct Scoring} (noCoT) and \textit{Pointwise CoT} under both Zero-Shot and SFT settings; under SFT, rationales are distilled from DeepSeek-R1 (DS) and Gemini-3-Pro (Gem). The best value is in \textbf{bold}. Dashes (--) indicate that Qwen3-0.6B fails to follow zero-shot noCoT instructions. Per-task BRIGHT scores are reported in Appendix~\ref{app:full_bright_results}.
}
\label{tab:main-results}
\end{table*}

\section{Introduction}


Integrating Chain-of-Thought (CoT) reasoning into Large Language Models (LLMs) for document reranking is a focal point in information retrieval~\citep{rank1,rankr1,tfrank}. However, in the pointwise paradigm, CoT models typically underperform direct scoring models. Recent works attribute this to inferior classification~\citep{Jedidi}, score polarization \citep{Jedidi,tfrank}, or calibration breakdown \citep{Overthinkingp}, prompting a critical question: can targeted training interventions bridge this performance gap?

We conduct a two-stage empirical study within a unified framework, utilizing Qwen series models (0.6B to 32B)~\citep{qwen2,qwen3} on different benchmarks. 
To verify that our conclusions are not specific to a single model family, we further replicate the core comparison on Llama-3.1-8B-Instruct~\citep{llama3}.
The first stage verifies the stability of this gap to rule out potential confounders. We scale model sizes and training data volumes under both zero-shot and supervised fine-tuning settings, and compare reasoning supervision distilled from DeepSeek-R1~\citep{deepseekr1} and Gemini-3-Pro~\citep{gemini3}. Results show that the relative gap between CoT and direct scoring remains stable across all settings.
This indicates that the disadvantage does not stem from model underfitting or insufficient reasoning quality, but rather has a more intrinsic source.
Building upon this, we design three stress tests to repair these potential defects: aligning classification boundaries via reinforcement learning, mitigating polarization through fine-grained supervision, and isolating generation from prediction using prompt decoupling. Experiments show that although these interventions successfully improve classification accuracy and absolute scores, the ranking performance gap persists. Our analysis suggests that routing continuous semantics through discrete text may constrain the resolution of ranking signals, revealing a stable bottleneck under current standard methods.


Our contributions are threefold: 1) we verify that the gap between pointwise CoT and direct scoring is robust to model scale, data volume, and supervision quality; 2) we design targeted stress tests that diagnose and mitigate its hypothesized causes; 3) we show that the gap persists despite absolute gains, suggesting that discrete text generation may impose an resolution constraint on ranking signals under current standard methods.
The code and data are available in the repository\footnote{\small\url{https://github.com/VerdureChen/Beyond-Polarization}}.

\begin{figure}[t]
  \centering
  \setlength{\belowcaptionskip}{-0.5cm}
  \includegraphics[width=\linewidth]{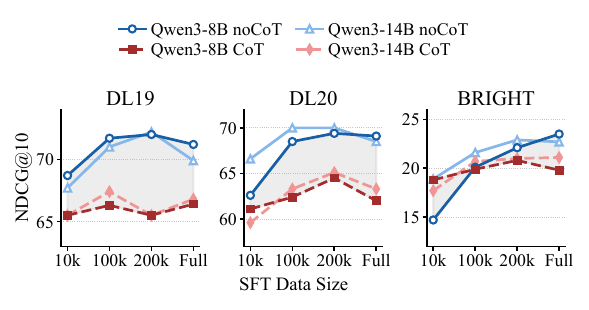}
  \vspace{-0.8cm} 
  \caption{
    Performance gap between Pointwise CoT and direct scoring (noCoT)
    across data scales on Qwen3. The Qwen2.5 results are provided in the Appendix~\ref{app:data_scaling_qwen25}.
  }
  \label{fig:data_scaling_gap}
\end{figure}

%% file: latex/experiment.tex
\section{Experimental Setup}
\label{sec:experiment}



 \textbf{Models and Strategies.} To systematically evaluate the pointwise reranking paradigm, we adopt the Qwen series~\citep{qwen2,qwen3} as our experimental backbone. This covers seven parameter scales across Qwen2.5 (7B, 14B, 32B) and Qwen3 (0.6B, 8B, 14B, 32B) to analyze scaling behaviors. 
 In addition, we include Llama-3.1-8B-Instruct~\citep{llama3} as a cross-family control to confirm that the observed gap generalizes beyond the Qwen series.
 We compare two mainstream pointwise variants: (1) \textbf{Direct Scoring (noCoT)}: Given an instruction $i$, a query $q$, and a passage $p$, the model directly predicts a binary relevance judgment formatted as \verb|<think>\n\n</think>|\allowbreak\verb|Answer: True/False|; (2) \textbf{Pointwise CoT}: The model auto-regressively generates an explicit reasoning chain before the final judgment, formatted as \verb|<think> Reasoning Trace </think> Answer:| \allowbreak \verb|True/False|. For both variants, the final relevance score for ranking is derived by weighting the logits of the candidate tokens at the decoding position immediately following the \verb|Answer:|.

 \textbf{Data Construction and Supervision.}
Based on the MS MARCO training set~\citep{msmarco} from Rank1~\citep{rank1}, we construct two variants of CoT training data: 
(1) \textbf{DS}: The original Rank1 set containing $\sim$380K samples, with reasoning chains distilled from DeepSeek-R1~\citep{deepseekr1}; 
(2) \textbf{Gem}: A variant replacing the original chains with those generated by Gemini-3-Pro~\citep{gemini3}. 
We prompt Gemini-3-Pro to simultaneously output a binary label and a 5-level fine-grained relevance score (0--4) for subsequent Fine-Grained Supervision experiments. Using the original binary labels for quality control, we filter out incorrect predictions, retaining $\sim$350K samples. These comparable data scales ensure a fair comparison between supervision sources during supervised fine-tuning.  We adopt Gemini-distilled rationales as the higher-quality supervision benchmark. See Appendix~\ref{app:data_quality} for a detailed quality comparison.
We evaluate on TREC DL19 and DL20~\citep{trec19,trec20} for standard fine-grained passage ranking, and BRIGHT~\citep{bright} for reasoning-intensive complex queries.
For more implementation details, see Appendix~\ref{app:implementation_details}.

%% file: latex/stage1.tex
\section{Stage I: Persistence of the Gap}\label{sec:persistence}

To rule out potential confounders such as model underfitting or suboptimal rationale quality, we systematically verify the robustness of the performance gap between Pointwise CoT and Direct Scoring (noCoT) across multiple dimensions.

\subsection{Robustness to Scale Expansion}


 \textbf{Insensitivity to Model Scaling.} 
Table~\ref{tab:main-results} compares seven model scales across the Qwen2.5 and Qwen3 series. While increasing model parameters significantly improves the absolute ranking ability of CoT models, it fails to achieve performance comparable to direct scoring. On the BRIGHT dataset, scaling Qwen3 from 0.6B to 32B boosts the absolute score of the SFT DS CoT model by over $2.4\times$ (from 9.6 to 23.0). However, regardless of model size, noCoT consistently maintains its lead. This relative disadvantage persists across different scales; on the DL19 dataset, Qwen3-32B exhibits the same inverted trend, with SFT noCoT scoring 73.7 against 70.1 for SFT DS CoT.

 \textbf{Insensitivity to Data Scaling.} 
To explore whether this performance inversion changes depending on the model's fitting stage to the training data, we conduct synchronous DS SFT experiments using \{10, 100, 200\}K, and the full instances on the 8B and 14B models. As shown in Figure~\ref{fig:data_scaling_gap},
experimental results demonstrate that this performance gap remains stable across the evaluated data scales. On BRIGHT, the model briefly relies on reasoning priors at the extremely low resource setting of 10K instances, where SFT CoT performs better on Qwen3-8B. However, as the data volume expands, SFT noCoT quickly overtakes SFT CoT. Subsequently, their performance curves rise in parallel, fixing the margin of disadvantage. On DL19 and DL20 datasets, the disparity is evident right from the 10K data mark.
This indicates that the lag of Pointwise CoT behind direct scoring is not an artifact of a specific fitting phase. Instead, it is an inherent paradigm limitation that persists throughout the entire training cycle, independent of both model and data scale.

\subsection{Robustness to Supervision Quality and Training Dynamics}
\label{subsec:supervision_robustness}

We further compare the impact of different supervision qualities and training dynamics under both zero-shot and SFT paradigms. As shown in Table~\ref{tab:main-results}, these comparisons yield two observations.

 \textbf{Insensitivity to Rationale Quality.} 
We replace the DeepSeek-R1 (DS) reasoning with ones distilled from Gemini-3-Pro, which features longer average chain lengths and more comprehensive evidence recall. However, this stronger teacher does not yield a fundamental breakthrough in SFT CoT ranking performance. On DL19 with Qwen3-14B, the Gemini-supervised CoT scores 65.7, slightly underperforming the DS variant at 66.8. 
This demonstrates that the performance gap cannot be simply attributed to suboptimal rationale quality.

 \textbf{Asymmetric Gains from SFT.} 
Interestingly, in the zero-shot setting without fine-tuning, small models typically rely on CoT to maintain basic performance. For instance, the zero-shot CoT performance of small models is comparable to or slightly better than noCoT. In extreme cases like Qwen3-0.6B, the model completely fails to follow instructions under the noCoT setting but successfully completes the task using CoT. In contrast, larger models like Qwen3-32B, equipped with stronger base capabilities, exhibit a clear noCoT advantage even in the zero-shot setting, achieving a 67.7 to 58.5 lead on DL19.
However, regardless of the initial state, SFT delivers more drastic gains for noCoT. For example, Qwen3-0.6B noCoT surges to 16.9, overtaking CoT at 9.6 on BRIGHT, while the existing lead of larger models is further consolidated. Although SFT comprehensively improves the performance of the models, it solidifies a margin of 1 to 4 points across all model scales.

%% file: latex/stage2.tex
\begin{table}[t]
\centering
\setlength{\tabcolsep}{3pt}
\setlength{\belowcaptionskip}{-0.4cm}
\resizebox{0.9\columnwidth}{!}{%
\begin{tabular}{@{}ll c ccc@{}}
\toprule
\multirow{2}{*}{\textbf{Bench.}} & \multirow{2}{*}{\textbf{Size}} 
& \multirow{2}{*}{\textbf{noCoT}}
& \multicolumn{3}{c}{\textbf{DeepSeek-R1 Distilled}~($\sim$380K)} \\
\cmidrule(l){4-6}
& & & \textbf{CoT} & \textbf{+GRPO} & \textbf{+Decouple} \\
\midrule
\multirow{3}{*}{DL19}
& 0.6B & 71.4 & 60.1\,\dneg{11.3} & 67.9\,\dneg{3.5} & 67.8\,\dneg{3.6} \\
& 8B   & 71.2 & 66.4\,\dneg{4.8}  & 70.3\,\dneg{0.9} & 67.4\,\dneg{3.8} \\
& 14B  & 69.9 & 66.8\,\dneg{3.1}  & 69.6\,\dneg{0.3} & 69.0\,\dneg{0.9} \\
\midrule
\multirow{3}{*}{DL20}
& 0.6B & 67.2 & 55.1\,\dneg{12.1} & 65.7\,\dneg{1.5} & 62.9\,\dneg{4.3} \\
& 8B   & 69.1 & 62.0\,\dneg{7.1}  & 66.9\,\dneg{2.2} & 61.2\,\dneg{7.9} \\
& 14B  & 68.5 & 63.3\,\dneg{5.2}  & 65.6\,\dneg{2.9} & 65.8\,\dneg{2.7} \\
\midrule
\multirow{3}{*}{BRIGHT}
& 0.6B & 16.9 & 9.6\,\dneg{7.3}   & 13.2\,\dneg{3.7} & 12.6\,\dneg{4.3} \\
& 8B   & 23.5 & 19.8\,\dneg{3.7}  & 20.8\,\dneg{2.7} & 21.7\,\dneg{1.8} \\
& 14B  & 22.7 & 21.1\,\dneg{1.6}  & 22.1\,\dneg{0.6} & 22.3\,\dneg{0.4} \\
\midrule\midrule
\multirow{2}{*}{\textbf{Bench.}} & \multirow{2}{*}{\textbf{Size}} 
& \multirow{2}{*}{\textbf{noCoT}}
& \multicolumn{3}{c}{\textbf{Gemini-3-Pro Distilled}~($\sim$350K)} \\
\cmidrule(l){4-6}
& & & \textbf{CoT} & \textbf{Mtag-noCoT} & \textbf{Mtag-CoT} \\
\midrule
\multirow{3}{*}{DL19}
& 0.6B & 71.4 & 62.5\,\dneg{8.9}  & \textbf{71.7}\,\dpos{0.3} & 65.9\,\dneg{5.5} \\
& 8B   & 71.2 & 64.2\,\dneg{7.0}  & \textbf{73.3}\,\dpos{2.1} & 68.9\,\dneg{2.3} \\
& 14B  & 69.9 & 65.7\,\dneg{4.2}  & \textbf{73.3}\,\dpos{3.4} & 70.7\,\dpos{0.8} \\
\midrule
\multirow{3}{*}{DL20}
& 0.6B & 67.2 & 55.7\,\dneg{11.5} & \textbf{68.7}\,\dpos{1.5} & 57.4\,\dneg{9.8} \\
& 8B   & 69.1 & 57.3\,\dneg{11.8} & \textbf{69.3}\,\dpos{0.2} & 67.3\,\dneg{1.8} \\
& 14B  & 68.5 & 61.7\,\dneg{6.8}  & \textbf{69.4}\,\dpos{0.9} & 66.1\,\dneg{2.4} \\
\midrule
\multirow{3}{*}{BRIGHT}
& 0.6B & 16.9 & 12.3\,\dneg{4.6}  & \textbf{18.0}\,\dpos{1.1} & 12.6\,\dneg{4.3} \\
& 8B   & 23.5 & 20.3\,\dneg{3.2}  & \textbf{24.1}\,\dpos{0.6} & 21.2\,\dneg{2.3} \\
& 14B  & 22.7 & 19.7\,\dneg{3.0}  & \textbf{25.1}\,\dpos{2.4} & 21.0\,\dneg{1.7} \\
\bottomrule
\end{tabular}%
}
\caption{NDCG@10 of stress-test interventions on TREC DL19, DL20, and BRIGHT. Subscripts show gaps to the baseline \textbf{noCoT}; per-row best in \textbf{bold}.}
\label{tab:stage2}
\end{table}

\begin{figure*}[t]
  \centering
  \setlength{\belowcaptionskip}{-0.4cm}
  \includegraphics[width=0.9\textwidth]{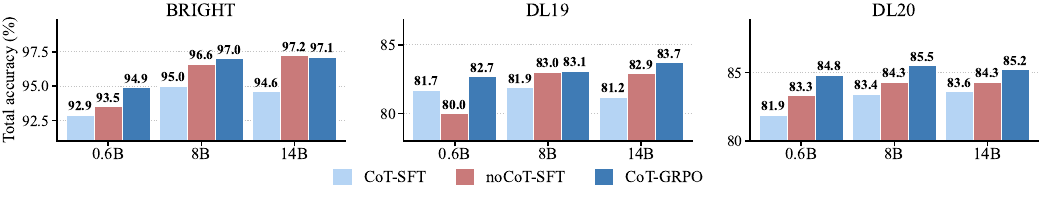}
\vspace{-0.4cm} 
  \caption{Total classification accuracy (\%) of Qwen3 across scales:
GRPO matches noCoT in 8/9 cells, yet NDCG@10 still trails
(Table~\ref{tab:stage2}).}

  \label{fig:grpo_cls}
  
\end{figure*}

\section{Stage II: Stress Tests on the Gap}
\label{sec:stage2}

To test whether targeted interventions can bridge this gap, we conduct three stress tests on the Qwen3 series (results in Table~\ref{tab:stage2}).

\textbf{Reinforcement Learning Alignment.}
Targeting inferior classification accuracy, we apply GRPO to SFT CoT to test whether explicitly aligning the classification boundary with Direct Scoring restores ranking performance. 
Given a generation $y$ and a binary label $g\!\in\!\{\textsc{true},\textsc{false}\}$, we integrate structural constraints as a gating function $\mathbf{1}[\text{valid}]$ (requiring paired \texttt{<think>...</think>} tags followed by an \texttt{Answer:\,<true|false>} pattern) directly into the reward function:
\begin{equation*} \resizebox{0.95\linewidth}{!}{$ R(y,g) = \mathbf{1}[\text{valid}] \cdot \big( \alpha\mathbf{1}[\text{thk}] + \beta\mathbf{1}[\text{fmt}] + \gamma\mathbf{1}[\text{match}] \big) $} \end{equation*} 
When valid, the reward is a weighted sum of formatting terms (structural validity $\alpha\!=\!0.1$, answer format $\beta\!=\!0.1$) and classification correctness, which dominates the gradient ($\gamma\!=\!0.8$). This design aligning the RL signal with the binary classification objective of Direct Scoring, allowing us to test: \textit{Does the ranking gap persist when both share the same classification boundary?}
As shown in Figure~\ref{fig:grpo_cls}, GRPO \textbf{matches or even surpasses} the classification accuracy of noCoT in most evaluation units. However, the corresponding NDCG@10 gap does not close synchronously: although the deficit for 14B on DL19 narrows from $-3.1$ to $-0.3$, noCoT still maintains a lead of $0.3$--$3.7$ across all units. \textit{Aligning classification boundaries does not fully translate into ranking quality, indicating a divergence between classification accuracy and ranking performance.}

\textbf{Fine-Grained Supervision.}
Targeting score polarization, we train \textbf{Mtag-noCoT} and \textbf{Mtag-CoT} using Gemini-3-Pro distilled data to test whether denser supervision improves the discrimination of partially relevant documents. After the \texttt{Answer:} prompt, both variants simultaneously output binary and fine-grained 0--4 scores, with the final relevance derived as an equal-weight ($1{:}1$) combination of a fine-grained score and a binary answer score, defined precisely in Appendix~\ref{app:score}.
Results show that Mtag-noCoT achieves the optimal performance. While Mtag-CoT improves significantly over standard Gemini CoT, it still lags behind Mtag-noCoT by $3.3$--$11.3$ NDCG. \textit{Fine-grained supervision synchronously raises the performance ceiling for both paradigms, failing to narrow the relative gap.}

\textbf{Architectural Decoupling.}
Targeting probability calibration breakdown caused by computational coupling within a single decoding pass, we design an explicit two-stage cascade. 
A generator $M_{g}$ first emits a rationale $t$, which is then concatenated with the original input and fed to a separate scorer $M_{s}$ that outputs only the score token.
Concretely, $M_s$ is \emph{not} trained from scratch: it is continually fine-tuned from the checkpoint of the trained generator $M_g$, sharing the identical output format and objective as noCoT (predicting True/False); the only difference is that its input additionally contains the pre-generated rationale $t$. Due to space limits, we instantiate this on the DS CoT data only.
This severs all computational coupling between reasoning and scoring while preserving the same training data and supervision.
As Table~\ref{tab:stage2} shows, \textbf{+Decouple} narrows the gap to standard CoT in most units, confirming coupling as a partial confounder. However, noCoT retains a 0.4 to 7.9 NDCG lead across all evaluations. \textit{Decoupling computation alleviates but does not eliminate the deficit; a residual gap remains because both stages still route their decision through the discrete reasoning trace $t$.}

\section{Analysis}
\label{sec:analysis}

In this section, we establish that the ranking gap is stable and generalizable, then probe its underlying mechanism and discuss the implications.

\textbf{Robustness to Random Seeds.}
To rule out single-run noise, we train Qwen3-14B under the noCoT and \textbf{+Decouple} settings with three random seeds. As shown in Table~\ref{tab:seed}, the systematic inter-group gap consistently exceeds intra-group variance across all benchmarks.

\begin{table}[t]
\centering
\small
\begin{tabular}{lccc}
\toprule
\textbf{Variant} & \textbf{DL19} & \textbf{DL20} & \textbf{BRIGHT} \\
\midrule
noCoT    & \textbf{71.90}{\scriptsize$\pm$1.74} & \textbf{69.33}{\scriptsize$\pm$0.76} & \textbf{23.50}{\scriptsize$\pm$0.69} \\
\textbf{+Decouple} & 69.00{\scriptsize$\pm$0.26} & 65.93{\scriptsize$\pm$0.47} & 22.33{\scriptsize$\pm$0.06} \\
\bottomrule
\end{tabular}
\caption{Mean\,$\pm$\,std of NDCG@10 for Qwen3-14B over three random seeds. The best value is in \textbf{bold}.}
\label{tab:seed}
\end{table}

\textbf{Generalization Beyond the Qwen Family.}
To verify the finding is not Qwen-specific, we replicate the core
comparison on Llama-3.1-8B-Instruct under identical settings. As reported
in Table~\ref{tab:llama}, noCoT again consistently outperforms CoT on the DL19/20 and BRIGHT datasets, showing that the gap generalizes across
model families rather than reflecting an idiosyncrasy of a single
backbone.

\begin{table}[t]
\centering
\small
\begin{tabular}{lccc}
\toprule
\textbf{Llama-3.1-8B-Instruct}  & \textbf{DL19} & \textbf{DL20} & \textbf{BRIGHT} \\
\midrule
noCoT & \textbf{73.7} & \textbf{69.8} & \textbf{19.8} \\
CoT   & 67.2 & 60.8 & 18.5 \\
\bottomrule
\end{tabular}
\caption{NDCG@10 on Llama-3.1-8B-Instruct. The best value is in \textbf{bold}.}
\label{tab:llama}
\end{table}

\textbf{Probing the Discrete-Text Bottleneck.}
We design a controlled probe to test whether the generated rationale
itself degrades the ranking signal. We train Qwen3-8B in CoT mode with the
label placed at \emph{both} the beginning and the end of the generated
sequence, keeping the rationale strictly in between. At inference the model
emits identical label text at both positions, yet ranking by their logits
diverges, as Table~\ref{tab:probe} reports. The end position, which follows
the rationale, drops markedly relative to the start, falling from $72.7$ to
$68.2$ on DL19 and from $22.0$ to $19.5$ on BRIGHT. Because the two
positions share an identical generation path and identical output text, this difference isolates the effect of the intervening rationale: passing the ranking signal through the generated text measurably attenuates it.

\begin{table}[t]
\centering
\small
\begin{tabular}{lccc}
\toprule
\textbf{Ranking position} & \textbf{DL19} & \textbf{DL20} & \textbf{BRIGHT} \\
\midrule
Start, before rationale & \textbf{72.7} & \textbf{69.2} & \textbf{22.0} \\
End, after rationale     & 68.2 & 66.0 & 19.5 \\
\bottomrule
\end{tabular}
\caption{NDCG@10 for the Label-CoT-Label probe when ranking by the label
before versus after the rationale on Qwen3-8B. The best value is in \textbf{bold}.}
\label{tab:probe}
\end{table}

\textbf{Discussion.}
We now offer a heuristic, empirical perspective on the gap that persists across our three stress tests. Let $h$ denote the continuous hidden state that predicts relevance and $t$ the discrete rationale that pointwise CoT must emit before scoring. Direct scoring reads relevance straight from $h$ and preserves its full resolution, whereas CoT first routes the signal through $t$. Because $t$ is a discrete sequence over a finite vocabulary, this routing imposes a data-processing-style bottleneck, whereby the ranking signal recoverable from $t$ cannot exceed what was already present in $h$, and the lossy nature of $t$ limits its resolution. Classification only needs the sign of the relevance signal at a threshold, yet ranking relies on the fine-grained residual to separate close pairs, and that residual is precisely what discretization into $t$ erodes.
This perspective unifies our three interventions. Fine-grained supervision densifies the signal that $t$ carries, Decouple severs the joint generation and scoring computation around $t$, and GRPO realigns the decision boundary defined over $t$. In every case $t$ stays on the critical path and caps the recoverable signal, most clearly for GRPO, whose classification accuracy matches or exceeds noCoT while NDCG still lags. The Label-CoT-Label probe of Table~\ref{tab:probe} makes this concrete. With identical label text and an identical generation path, the ranking signal still drops once it is routed through the intervening rationale.

\section{Conclusion}
\label{sec:conclusion}

In this paper, we investigate the performance gap between pointwise CoT and direct scoring from multiple dimensions.
Our study indicates that the underperformance of pointwise CoT relative to direct scoring in reranking remains stable across model capacity, data scale, and reasoning quality. Targeted stress tests via reinforcement learning alignment, fine-grained supervision, and architectural decoupling alleviate specific deviations but fail to eliminate the ranking deficit. 
This suggests that, specifically within the pointwise scoring paradigm, routing continuous semantics through discrete text constrains the resolution of ranking signals. We frame this as a bottleneck that is stable and difficult to overcome under existing standard methods, while explicitly leaving open that novel architectures or training strategies may further narrow the gap. 

%% file: latex/appendix.tex
\section{Related Work}
\label{app:related_work}




\paragraph{Integration of Reasoning into Reranking.}
LLM-based reranking is typically categorized into pointwise~\citep{rank1,tfrank}, pairwise~\citep{prp}, and listwise paradigms~\citep{rankgpt,rankk,rearank}. With the advent of Large Reasoning Models (LRMs) such as DeepSeek-R1~\citep{deepseekr1}, OpenAI's o-series~\citep{gpt4,gpt5}, and the Qwen3 series~\citep{qwen3}, researchers have begun integrating explicit reasoning processes into reranking pipelines, aiming to enhance discriminative capabilities on reasoning-intensive queries \citep{reasoning_survey}. For instance, Rank1~\citep{rank1} systematically distills R1-style reasoning into pointwise rerankers, enabling smaller models to produce explicitly explainable relevance judgments while preserving pointwise parallelizability. TFRank~\citep{tfrank} further explores a practical deployment path for small-scale pointwise reasoning rerankers: the model learns to reason during training but directly outputs scores in a ``think-free'' manner during inference, thereby maintaining performance while significantly reducing latency.
Other studies have predominantly explored reasoning within listwise and setwise settings. Rank-K \citep{rankk} leverages the test-time compute of LRMs for listwise reranking, improving performance on hard queries through extended thinking processes and demonstrating strong multilingual transferability. REARANK \citep{rearank} trains a listwise reranking model as an explicit reasoning agent using only 179 annotated samples and reinforcement learning. Under a setwise prompting framework, Rank-R1 \citep{rankr1} elicits the reasoning capabilities of rerankers using relatively sparse relevance labels and the GRPO algorithm, exhibiting robust generalization on complex out-of-domain queries. Similarly, ReasonRank \citep{ReasonRank} proposes an automated framework for synthesizing reasoning-heavy training data and employs a two-stage training strategy, significantly outperforming existing baselines in listwise reasoning reranking. Focusing on distillation and interpretability, ReasoningRank \citep{reasoningrank} simultaneously distills ``explicit reasoning'' and ``comparative reasoning'' from a teacher LLM to a student model.
Beyond model architecture and training strategies, recent efforts also advance reasoning-intensive retrieval from the perspectives of data and computational efficiency. Addressing the prevalence of short, factual queries in existing retrieval corpora, ReasonIR \citep{ReasonIR} constructs a synthetic data generation pipeline tailored for general reasoning tasks. Conversely, LIMRANK \citep{lessismore} emphasizes a ``less is more'' training paradigm, constructing high-quality, diverse, and realistic small-scale training data for reranking.

\paragraph{Diagnostic Perspectives on the Pointwise CoT Gap.}
Despite the general success of explicit reasoning in listwise settings, several studies have observed a counterintuitive phenomenon under the pointwise paradigm: Chain-of-Thought unexpectedly underperforms direct scoring. Recent works have attempted to diagnose this issue. Under strictly controlled training conditions, \citet{Jedidi} systematically compared models with and without reasoning. They found that models retaining reasoning performed worse and exhibited lower binary classification accuracy than direct scoring models. They attributed this to the reasoning process pushing relevance probabilities toward extremes (i.e., score polarization), thereby impairing the model's ability to discriminate ``partially relevant'' documents at a fine-grained level, which also echoed by TFRank \citep{tfrank}. Similarly, \citet{Overthinkingp} discovered that reasoning in pointwise reranking breaks probability calibration. Furthermore, they noted that while reasoning improves in-domain fitting in listwise reranking, it introduces higher variance and poorer out-of-domain generalization.
Existing diagnostic studies primarily offer phenomenon-level explanations. Taking these as our starting point, we instantiate targeted repairs for these diagnostics into three systematic stress tests: applying GRPO to align classification boundaries, introducing fine-grained multi-label supervision to mitigate score polarization, and adopting a two-stage cascade architecture to eliminate generation-prediction coupling. As we systematically report, while each intervention successfully mitigates its targeted deviation, the relative ranking gap between CoT and direct scoring persistently remains across nine evaluation units. Based on this robust empirical evidence, we unify these diverse diagnostics as different statistical manifestations of the same ``discrete text bottleneck,'' providing a comprehensive empirical characterization of this inherent limitation.

\section{Comparison of Distillation Data Quality}
\label{app:data_quality}

In Section~\ref{sec:persistence}, we explore whether rationale quality influences the performance gap by comparing rationales distilled from DeepSeek-R1 and Gemini-3-Pro. This section provides detailed statistics, prompt designs, and case studies to support our assessment that the Gemini-distilled data exhibits higher reasoning quality under human evaluation.

\subsection{Data Statistics and Characteristics}

Table~\ref{tab:data_comparison} presents the macro-level statistics of the two distilled datasets. While the DeepSeek-R1 distilled data adopts a highly conversational, stream-of-consciousness style (e.g., "Okay, let's see..."), the Gemini-3-Pro distilled data enforces a strict, structured reasoning paradigm. On average, the Gemini rationales are 30\% longer (1489 vs. 1151 characters), providing denser analytical steps before arriving at the final binary decision. 

\begin{table}[h]
\centering
\resizebox{\linewidth}{!}{%
\begin{tabular}{lcc}
\toprule
\textbf{Metric} & \textbf{DeepSeek-R1 (DS)} & \textbf{Gemini-3-Pro (Gem)} \\
\midrule
Size & $\sim$385k (760MB) & $\sim$352k (831MB) \\
Style & Conversational & Structured w/ bullets \\
Avg. Length & $\sim$1151 chars & $\sim$1489 chars (+30\%) \\
Output Format & \texttt{<think>...</think>} & \texttt{<think>...</think>} \\
Score Type & Binary only & Fine-grained (0-4) + Binary \\
\bottomrule
\end{tabular}%
}
\caption{Comparison of distilled datasets.}
\label{tab:data_comparison}
\end{table}

\subsection{Distillation Prompt for Gemini-3-Pro}

To explicitly guide the teacher model to produce high-quality rationales and mitigate score polarization, we prompt Gemini-3-Pro to simultaneously output a fine-grained relevance score (0--4) and a binary label. The core instructions are as follows:

\begin{tcolorbox}[colback=gray!5!white, colframe=gray!50!black, arc=2mm, boxrule=0.5pt, left=6pt, right=6pt, top=4pt, bottom=4pt]
\small
\textbf{Instruction for Gemini-3-Pro:} \\
For EVERY passage listed above, provide a rigorous step-by-step analysis:
\begin{enumerate}[label=\arabic*., itemsep=0pt, parsep=0pt, topsep=4pt, leftmargin=*]
    \item \textbf{Query Analysis}: Identify the user's core intent, key constraints, and required entity types.
    \item \textbf{Per-Passage Reasoning}: For each passage, analyze:
    \begin{itemize}[itemsep=0pt, parsep=0pt, topsep=4pt, leftmargin=12pt]
        \item \textbf{Helpfulness}: Does this passage help answer the query?
        \item \textbf{Directness}: Is the information direct or requires excessive inference?
        \item \textbf{Matching}: Entities, time, and scope alignment.
    \end{itemize}
    \item \textbf{Scoring \& Decision}: Assign a score (0-4) and binary label per passage.
\end{enumerate}
\end{tcolorbox}

In Table~\ref{tab:main-results}, we utilize \textbf{only the binary output} from this dataset to ensure fair comparison under the binary setting. In the fine-grained supervision experiments (Section~\ref{tab:stage2}), we utilize \textbf{both the 0--4 labels and binary outputs}. To empirically validate the quality difference, we conducted a manual human evaluation on 200 randomly sampled query-document pairs. Human annotators preferred Gemini-distilled rationales over DeepSeek-R1 rationales in \textbf{70\%} of the cases (win rate), confirming the superiority of the structured generation.

\subsection{Case Study: Reasoning Styles}

Table~\ref{tab:case_study} provides a concrete example illustrating the significant difference in reasoning styles when evaluating a highly confusable negative sample. The user queries the approval rate of disability claims \textit{after a hearing}, while the retrieved document provides the approval rate for \textit{initial} applications.

The DeepSeek-R1 (DS) rationale evaluates relevance through a conversational, stream-of-consciousness narrative (e.g., ``Wait, the query is specifically about... I know that...''). It successfully identifies the mismatch but does so by meandering through external knowledge before arriving at the conclusion.

In contrast, the Gemini rationale strictly adheres to the requested analytical paradigm. It systematically extracts the exact statistic requested in Step 1, and the provided statistic in Step 2. Most importantly, in Step 3, it explicitly decomposes the evaluation into \textit{Topic Match} and \textit{Aspect Mismatch}. It sharply defines the boundary: while both deal with disability claims, ``initial application vs. hearing'' represents two entirely different stages in the appeals process. This level of structural rigor and zero-shot aspect-level discrimination makes the Gemini dataset highly aligned with human preference for analytical reasoning, making it a substantially stronger teacher for supervised fine-tuning.

\begin{table*}[t]
\centering
\small
\resizebox{\textwidth}{!}{%
\begin{tabular}{p{0.15\linewidth} | p{0.8\linewidth}}
\toprule
\textbf{Query} & percent of disability claims approved after hearing \\
\textbf{Document} & How Often Does Social Security Approve Disability The First Time You Apply? Statistics indicate that the national average for approval on initial disability claims is about thirty to thirty-five percent... \\
\midrule
\textbf{DeepSeek-R1} & 
\texttt{<think>} Okay, let's see. The user is asking about the percentage of disability claims approved after a hearing. The passage provided talks about approval rates on initial disability claims... Hmm. \newline
Wait, the query is specifically about claims approved "after a hearing." [...] I know that the Social Security disability process usually has multiple stages: initial application, reconsideration, and then a hearing... The initial approval rate is different from the approval rate after a hearing. So the passage is about the first time you apply... Therefore, the passage isn't directly relevant. \texttt{</think>} \textbf{Answer: false} \\
\midrule
\textbf{Gemini-3-Pro} & 
\texttt{<think>}
\textbf{1. Query Analysis}: The user is asking for a specific statistic: the percentage of disability claims that are approved at the "hearing" stage... \newline
\textbf{2. Passage Analysis}: The passage provides statistics for the \textit{initial} disability claim ("The First Time You Apply"). It states the national average for initial approval is about 30-35\%. \newline
\textbf{3. Relevance \& Utility Analysis}: \newline
\indent $\bullet$ \textit{Topic Match}: Both deal with Social Security disability claim approval rates. \newline
\indent $\bullet$ \textit{Aspect Mismatch}: The query specifically asks for approval rates \textit{after a hearing}. The passage specifically provides approval rates for \textit{initial claims}. These are two completely different stages... \newline
\indent $\bullet$ \textit{Utility}: The passage does not answer the question. Knowing the initial approval rate does not tell the user what the approval rate is at the hearing level. \newline
\textbf{4. Scoring \& Decision}: The passage is on the correct general topic but answers a different specific question. This fits the definition of a Score 1 (Marginal/Related topic but wrong aspect). \newline
\texttt{</think>} \textbf{Answer: false} \\
\bottomrule
\end{tabular}%
}
\caption{Case study comparing the conversational reasoning of DeepSeek-R1 versus the structured reasoning of Gemini-3-Pro on a challenging negative sample. Text is slightly truncated for brevity.}
\label{tab:case_study}
\end{table*}

\section{Implementation Details and Prompt Templates}
\label{app:implementation_details}

\subsection{Training Details}

Our experiments are conducted on a cluster of four NVIDIA H20 GPUs.

\paragraph{SFT.}
All supervised fine-tuning experiments are implemented based on the LLaMA-Factory framework, employing a full-parameter fine-tuning strategy. To ensure that the models correctly learn and output structured reasoning chains, we explicitly append \texttt{<think>} and \texttt{</think>} as special tokens to the tokenizer vocabulary for the Qwen2.5 series models. Regarding hyperparameters, the models are trained using bfloat16 precision with a learning rate of $1.0 \times 10^{-5}$ governed by a cosine learning rate scheduler. The warmup ratio is set to $0.05$, and the training is conducted for $1.0$ epoch.

\paragraph{GRPO.}
For the GRPO experiments, we employ a sampling strategy to curate appropriate training data. Specifically, we perform inference 8 times on the training set using the SFT-tuned models, and selectively retain samples that are answered between 2 and 6 times. 
The reinforcement learning interventions are implemented using the \texttt{verl} framework. To ensure the model maintains its foundational reasoning formats and prior capabilities during exploration, we initialize the policy model with the checkpoint obtained from the aforementioned CoT SFT phase. The core hyperparameters are configured as follows: a learning rate of $1.0 \times 10^{-6}$ with a global training batch size of $256$, trained over $2$ epochs. Sequence length constraints are set with a maximum prompt length of $2048$ and a maximum response length of $4096$.

\subsection{Final Score Computation for Mtag}
\label{app:score}
The composite score $s_{\text{cot}}$ combines two parts.
\textbf{(i) Fine-grained score.} We read the \texttt{<score>} tag over the
five candidate tokens $\{0,1,2,3,4\}$ and apply a softmax over their
logits $z_k$ to obtain the probabilities $p_k$, then take the expectation
and normalize:
\begin{equation}
s_{\text{score}}=\frac{1}{4}\sum_{k=0}^{4} k\, p_k,\qquad
p_k=\mathrm{softmax}(z_k),
\end{equation}
where $z_k$ is the logit assigned by the generator to the $k$-th
relevance-level token at the \texttt{<score>} position.
\textbf{(ii) Answer score.} We read the \texttt{<answer>} tag and
normalize the True probability against True and False:
\begin{equation}
s_{\text{ans}}=\frac{p_{\text{true}}}{p_{\text{true}}+p_{\text{false}}}.
\end{equation}
\textbf{Final.} The two are averaged with equal weight:
\begin{equation}
s_{\text{cot}}=\tfrac{1}{2}\,s_{\text{score}}+\tfrac{1}{2}\,s_{\text{ans}}.
\end{equation}

\subsection{Prompt Templates}
\label{app:prompt_templates}

We design strict prompt templates for different paradigms to ensure consistency in model output formats during both training data construction and inference evaluation. The complete templates for the three core paradigms are provided below.

\vspace{1em}
\noindent\textbf{1. Direct Scoring (noCoT)} \\
This template requires the model to directly output a binary boolean value, bypassing any intermediate reasoning steps.

\begin{tcolorbox}[colback=gray!5!white, colframe=gray!50!black, arc=2mm, boxrule=0.5pt, left=6pt, right=6pt, top=4pt, bottom=4pt]
\small
Assess the relevance of the query to the document using a binary scale (true or false):

* true: relevant - the document directly addresses the query or provides necessary information.

* false: not relevant - the document is unrelated.

Write 'Answer: ' followed by only your rating (true or false).

Query: \texttt{\{\{query\}\}} \\
Document: \texttt{\{\{document\}\}}
\end{tcolorbox}

\vspace{1em}
\noindent\textbf{2. Pointwise CoT} \\
This template prompts the model to generate a discrete natural language reasoning process before outputting the final binary score.

\begin{tcolorbox}[colback=gray!5!white, colframe=gray!50!black, arc=2mm, boxrule=0.5pt, left=6pt, right=6pt, top=4pt, bottom=4pt]
\small
Assess the relevance of the query to the document using a binary scale (true or false):

* true: relevant - the document directly addresses the query or provides necessary information.

* false: not relevant - the document is unrelated.

First, provide a concise reason for your rating. Then, on a new line, write 'Answer: ' followed by only your rating (true or false).

Query: \texttt{\{\{query\}\}} \\
Document: \texttt{\{\{document\}\}}
\end{tcolorbox}

\vspace{1em}
\noindent\textbf{3. Fine-Grained Supervision (Mtag)} \\
This template introduces a continuous semantic signal (0--4 scale), requiring the model to complete reasoning within the \texttt{<think>} tags and strictly output a structured combination of the fine-grained score and the binary result.

\begin{tcolorbox}[colback=gray!5!white, colframe=gray!50!black, arc=2mm, boxrule=0.5pt, left=6pt, right=6pt, top=4pt, bottom=4pt]
\small
You are a relevance evaluator. Your task is to determine if the following document is relevant to the query. \\
Please analyze the relationship and provide a relevance score from 0 to 4, based on the following scale: \\
- 0: Completely irrelevant \\
- 1: Mostly irrelevant \\
- 2: Partially relevant \\
- 3: Relevant \\
- 4: Highly relevant

Format your response strictly as follows: \\
\texttt{<think>} \\
Your reasoning process here... \\
\texttt{</think>} \\
\texttt{<score>}0-4\texttt{</score>} \\
\texttt{<answer>}true/false\texttt{</answer>}

Query: \texttt{\{\{query\}\}} \\
Document: \texttt{\{\{document\}\}}
\end{tcolorbox}

\section{Data Scaling Results on Qwen2.5 Series}
\label{app:data_scaling_qwen25}
\begin{figure}[t] \centering \includegraphics[width=\columnwidth]{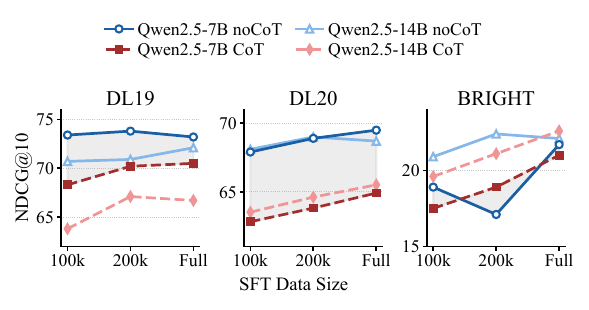} \caption{Data scaling on Qwen2.5 series.} \label{fig:scaling-qwen25} \end{figure}

See Figure~\ref{fig:scaling-qwen25} for the data scaling results on the Qwen2.5 series models. The trends are consistent with those observed in the main paper, further confirming the persistence of the CoT gap across different model series and sizes.

\section{Full Results of BRIGHT Benchmark}
\label{app:full_bright_results}

See Table~\ref{tab:bright_results} for the complete reranking performance on the BRIGHT benchmark across all evaluated models and methods. The results are reported in nDCG@10 for each of the 12 evaluation units, along with the average (AVG) across all units. This comprehensive table allows for a detailed comparison of the impact of different methods on both Qwen2.5 and Qwen3 series models of varying sizes.

\begin{table*}[t]
\centering
\small
\setlength{\tabcolsep}{3pt}
\renewcommand{\arraystretch}{1.05}
\begin{tabular}{llccccccccccccc}
\toprule
\textbf{Model} & \textbf{Method} & \textbf{Bio.} & \textbf{Earth.} & \textbf{Econ.} & \textbf{Psy.} & \textbf{Rob.} & \textbf{Stack.} & \textbf{Sus.} & \textbf{Leet.} & \textbf{Pony} & \textbf{AoPS} & \textbf{TQ-Q} & \textbf{TQ-T} & \textbf{AVG} \\
\midrule
\multicolumn{15}{c}{\textit{Qwen2.5}} \\
\midrule
\multirow{5}{*}{7B}
 & Zero-Shot noCoT & 15.9 & 25.3 & 10.0 & 16.8 & 13.5 & 14.0 & 11.6 &  9.6 &  7.6 & 1.3 &  1.0 &  6.9 & 11.1 \\
 & Zero-Shot CoT   & 20.5 & 33.5 & 15.3 & 21.9 & 17.4 & 17.3 & 16.9 & 11.2 &  9.2 & 3.9 &  6.8 &  9.2 & 15.3 \\
 & SFT noCoT       & 34.0 & 42.3 & 23.4 & 30.6 & 21.1 & 24.9 & 26.5 & 18.2 & 20.2 & 1.7 &  7.0 & 10.9 & 21.7 \\
 & SFT DS          & 32.1 & 39.5 & 21.2 & 28.2 & 19.0 & 22.8 & 25.2 & 18.9 & 18.8 & 6.5 &  8.7 & 11.1 & 21.0 \\
 & SFT Gem         & 28.5 & 35.8 & 19.1 & 24.6 & 18.4 & 22.1 & 22.6 &  9.1 & 14.1 & 6.5 &  6.6 &  8.3 & 18.0 \\
\midrule
\multirow{5}{*}{14B}
 & Zero-Shot noCoT & 29.4 & 39.7 & 22.5 & 25.8 & 21.6 & 22.7 & 23.9 & 16.7 & 11.0 & 4.3 &  8.1 & 10.4 & 19.7 \\
 & Zero-Shot CoT   & 29.5 & 36.2 & 17.1 & 23.9 & 19.8 & 20.8 & 24.5 & 11.8 &  7.7 & 5.1 &  6.1 & 10.5 & 17.8 \\
 & SFT noCoT       & 31.7 & 43.6 & 25.9 & 30.4 & 22.8 & 24.5 & 27.6 & 18.0 & 14.6 & 5.9 &  8.9 & 10.9 & 22.1 \\
 & SFT DS          & 31.0 & 41.7 & 25.4 & 28.5 & 22.3 & 24.4 & 26.3 & 22.6 & 21.3 & 7.3 &  9.9 & 10.7 & 22.6 \\
 & SFT Gem         & 29.0 & 40.5 & 22.9 & 26.8 & 22.5 & 23.6 & 25.0 & 11.1 &  9.5 & 2.3 &  6.1 &  6.6 & 18.8 \\
\midrule
\multirow{5}{*}{32B}
 & Zero-Shot noCoT & 28.9 & 41.3 & 23.6 & 28.9 & 23.0 & 24.5 & 24.6 & 19.0 & 11.4 & 9.4 &  9.2 & 11.7 & 21.3 \\
 & Zero-Shot CoT   & 29.9 & 39.6 & 19.0 & 26.9 & 19.2 & 19.0 & 24.2 &  8.1 &  8.8 & 4.3 &  4.8 &  9.0 & 17.7 \\
 & SFT noCoT       & 32.8 & 47.8 & 27.1 & 32.7 & 23.4 & 25.8 & 30.3 & 23.1 & 17.8 & 10.5 &  9.9 & 11.3 & 24.4 \\
 & SFT DS          & 33.2 & 44.1 & 23.7 & 30.0 & 18.9 & 23.2 & 25.9 & 19.9 & 21.2 & 9.3 & 10.7 & 11.1 & 22.6 \\
 & SFT Gem         & 29.0 & 40.0 & 25.1 & 27.0 & 23.2 & 24.4 & 23.6 & 12.2 & 11.3 & 4.1 &  5.2 &  9.4 & 19.5 \\
\midrule
\multicolumn{15}{c}{\textit{Qwen3}} \\
\midrule
\multirow{5}{*}{0.6B}
 & Zero-Shot noCoT & --   & --   & --   & --   & --   & --   & --   & --   & --   & --  & --   & --   & --   \\
 & Zero-Shot CoT   & 13.5 & 15.7 & 10.7 &  9.3 &  7.2 &  8.8 & 10.2 & 16.9 &  5.1 & 4.2 &  4.6 &  4.4 &  9.2 \\
 & SFT noCoT       & 19.4 & 30.3 & 14.1 & 23.4 & 14.4 & 18.0 & 17.4 & 24.2 & 18.8 & 4.1 &  8.6 &  9.6 & 16.9 \\
 & SFT DS          & 15.2 & 20.1 &  8.6 & 13.4 &  5.8 &  8.9 & 12.6 & 11.9 &  3.8 & 3.3 &  5.0 &  7.0 &  9.6 \\
 & SFT Gem         & 21.2 & 27.2 & 12.0 & 18.7 &  9.7 & 14.2 & 16.6 & 13.5 &  2.5 & 2.9 &  6.5 &  3.0 & 12.3 \\
\midrule
\multirow{5}{*}{8B}
 & Zero-Shot noCoT & 28.1 & 42.0 & 22.5 & 24.1 & 14.9 & 19.6 & 21.2 & 24.3 & 11.2 & 8.0 & 11.0 & 11.0 & 19.8 \\
 & Zero-Shot CoT   & 25.0 & 36.8 & 16.8 & 21.9 & 17.7 & 19.3 & 18.4 & 11.6 &  3.3 & 6.0 &  6.6 &  9.7 & 16.1 \\
 & SFT noCoT       & 31.4 & 44.3 & 23.1 & 30.3 & 22.4 & 25.2 & 27.0 & 27.0 & 23.0 & 6.3 & 11.3 & 11.0 & 23.5 \\
 & SFT DS          & 30.9 & 38.9 & 20.3 & 24.7 & 16.3 & 22.1 & 23.4 & 17.3 & 17.6 & 6.4 &  8.9 & 10.9 & 19.8 \\
 & SFT Gem         & 27.4 & 41.4 & 24.4 & 28.9 & 21.7 & 23.5 & 26.4 & 11.6 & 17.6 & 3.5 &  8.0 &  9.8 & 20.3 \\
\midrule
\multirow{5}{*}{14B}
 & Zero-Shot noCoT & 28.8 & 40.5 & 23.5 & 27.1 & 22.4 & 21.7 & 23.4 & 27.8 &  8.6 & 9.2 & 10.1 & 11.5 & 21.2 \\
 & Zero-Shot CoT   & 28.5 & 37.7 & 23.1 & 26.3 & 19.3 & 19.6 & 22.3 & 14.4 &  9.7 & 6.4 &  9.1 &  9.7 & 18.8 \\
 & SFT noCoT       & 29.8 & 42.6 & 23.6 & 28.9 & 23.8 & 24.4 & 24.9 & 28.7 & 14.9 & 7.6 & 11.6 & 11.5 & 22.7 \\
 & SFT DS          & 28.9 & 40.5 & 23.6 & 25.2 & 18.6 & 21.1 & 23.7 & 22.7 & 21.9 & 7.8 &  8.8 & 10.8 & 21.1 \\
 & SFT Gem         & 31.7 & 41.8 & 22.7 & 28.9 & 24.3 & 24.1 & 26.4 &  4.5 & 15.4 & 1.7 &  4.4 & 10.9 & 19.7 \\
\midrule
\multirow{5}{*}{32B}
 & Zero-Shot noCoT & 30.8 & 43.0 & 25.9 & 30.1 & 23.7 & 24.4 & 26.2 & 22.8 &  7.9 & 11.2 & 10.2 & 12.4 & 22.4 \\
 & Zero-Shot CoT   & 26.4 & 36.0 & 22.3 & 26.7 & 18.8 & 19.5 & 20.9 &  7.1 & 10.0 & 3.4 &  6.9 &  8.8 & 17.2 \\
 & SFT noCoT       & 32.5 & 45.5 & 27.5 & 32.0 & 23.2 & 28.0 & 28.9 & 27.8 & 21.3 & 5.9 & 11.0 & 11.6 & 24.6 \\
 & SFT DS          & 35.1 & 44.5 & 26.2 & 28.4 & 20.6 & 24.3 & 26.9 & 19.1 & 21.0 & 7.2 & 11.1 & 11.3 & 23.0 \\
 & SFT Gem         & 34.1 & 41.9 & 26.7 & 28.6 & 26.9 & 26.9 & 28.6 & 19.1 &  8.8 & 2.5 &  8.5 &  9.5 & 21.8 \\
\bottomrule
\end{tabular}
\caption{Per-task NDCG@10 of pointwise reranking on the BRIGHT benchmark across the Qwen2.5 and Qwen3 series, complementing the aggregated results in the main text. We compare \textit{Direct Scoring} (noCoT) and \textit{Pointwise CoT} under both Zero-Shot and Supervised Fine-Tuning (SFT) settings; under SFT, rationales are distilled from DeepSeek-R1 (DS) and Gemini-3-Pro (Gem). The horizontal rule separates Zero-Shot from SFT regimes within each model. Within each (setting, model) block, the best value per column is in \textbf{bold}. Dashes (--) indicate that Qwen3-0.6B fails to follow zero-shot noCoT instructions.}
\label{tab:bright_results}
\end{table*}